\documentclass[runningheads]{llncs}
\usepackage[T1]{fontenc}
\usepackage{graphicx}
\usepackage{amsmath}
\usepackage{amsfonts}
\begin{document}
\title{Cross-Domain Evaluation of Few-Shot Classification Models: Natural Images vs. Histopathological Images}
\titlerunning{Cross-Domain Few-shot}
%
\author{Ardhendu Sekhar \and Aditya Bhattacharya \and Vinayak Goyal \and Vrinda Goel \and Aditya Bhangale \and Ravi Kant Gupta \and Amit Sethi}
\authorrunning{Ardhendu et al.}
%
\institute{Department of Electrical Engineering, Indian Institute of Technology Bombay, Mumbai, India}
\maketitle              
\begin{abstract}
In this study, we investigate the performance of few-shot classification models across different domains, specifically natural images and histopathological images. We first train several few-shot classification models on natural images and evaluate their performance on histopathological images. Subsequently, we train the same models on histopathological images and compare their performance. We incorporated four histopathology datasets and one natural images dataset and assessed performance across 5-way 1-shot, 5-way 5-shot, and 5-way 10-shot scenarios using a selection of state-of-the-art classification techniques. Our experimental results reveal insights into the transferability and generalization capabilities of few-shot classification models between diverse image domains. We analyze the strengths and limitations of these models in adapting to new domains and provide recommendations for optimizing their performance in cross-domain scenarios. This research contributes to advancing our understanding of few-shot learning in the context of image classification across diverse domains.

\end{abstract}
\section{Introduction}
Few-shot learning has emerged as a powerful paradigm for addressing the challenges of classification tasks in scenarios where labeled data is scarce. By leveraging a small number of annotated examples, few-shot classification models can generalize effectively to new classes or domains. This capability holds significant promise for various real-world applications, particularly in medical imaging where labeled data is often limited and costly to obtain. However, the effectiveness of few-shot learning across diverse domains remains a topic of ongoing investigation. In this study, we explore the transferability and generalization capabilities of few-shot classification models across different domains, focusing specifically on natural images and histopathological images. Natural images, representative of everyday scenes and objects, present diverse visual characteristics, while histopathological images depict tissue samples at a microscopic level, typically used for medical diagnoses. Understanding how few-shot classification models perform when transitioning between these distinct domains is crucial for assessing their practical utility in real-world applications. Our investigation involves training several state-of-the-art few-shot classification models on natural image datasets and evaluating their performance on histopathological images. We then repeat the process in the reverse direction, training the same models on histopathological images and testing them on both natural and histopathological image datasets. By comparing the performance of these models across different domains, we aim to uncover insights into their adaptability and robustness in cross-domain classification tasks.

In this study, we conducted evaluations of several state-of-the-art few-shot classification techniques using histopathology medical datasets and a natural image dataset. Specifically, we utilized a dataset prepared by Komura et al.\cite{1.6M}, FHIST\cite{FHIST} and Mini-Imagenet\cite{miniimagenet} for our experiments. Mini-Imagenet dataset is a standard natural image dataset meant for few-shot classification techniques. The FHIST dataset encompasses multiple histology datasets, including CRC-TP\cite{crctp}, NCT-CRC-HE-100K\cite{NCT}, LC25000\cite{lc25000}, and BreakHis\cite{BreakHis}. For our experiments, we focused on CRC-TP, NCT-CRC-HE-100K, and LC25000 datasets. CRC-TP is a colon cancer dataset with six classes, while NCT-CRC-HE-100K is another colon cancer dataset with nine classes. LC25000 contains both lung and colon cancer images, featuring five classes. Additionally, we utilized a histology dataset introduced by Komura et al.\cite{1.6M}, which comprises approximately 1.6 million cancerous image patches from 32 different organs in the body. The classes in this dataset are categorized based on distinct organ sites. We employed the datasets by Komura et al.\cite{1.6M} adn miniimagenet to train our few-shot classification models, which were subsequently evaluated on various FHIST datasets.

In our thorough investigation of few-shot classification methodologies, we have systematically integrated a range of state-of-the-art techniques to ensure a comprehensive evaluation. The methodologies employed in our experiments encompass Prototypical Networks\cite{Prototypical}, SimpleShot\cite{Simpleshot}, LaplacianShot\cite{Laplacianshot}, DeepEMD\cite{DeepEMD} and DeepBDC\cite{DeepBDC}. Prototypical Networks\cite{Prototypical}, a cornerstone technique in few-shot learning, utilize prototypes as representative embeddings for each class. By minimizing the distance between query examples and class prototypes, this model excels in quickly adapting to new classes with limited labeled data. SimpleShot\cite{Simpleshot} emphasizes simplicity in few-shot learning, relying on straightforward yet effective techniques. This highlights the efficacy of simplicity in addressing complex classification tasks with limited data. LaplacianShot\cite{Laplacianshot} introduces Laplacian regularization to enhance few-shot learning performance. By integrating this regularization technique, the model aims to enhance generalization and robustness across diverse classes. DeepEMD\cite{DeepEMD} learns image representations by quantifying the discrepancy between joint characteristics of embedded features and the product of the marginals, facilitating comprehensive understanding of image similarities. DeepBDC\cite{DeepBDC}, an advanced version of prototypical networks which uses a Brownian Distance Covriance matrix, realised in the form of a CNN module, calculates the prototypes of the classes in episodes of  few-shot setting. Then the distance between the query examples and the support prototypes is minimised.

This research contributes to advancing the understanding of few-shot learning algorithms and their applicability in diverse image classification scenarios. The findings from our study hold implications for various fields, including medical imaging, where the ability to transfer knowledge across domains can significantly enhance diagnostic accuracy and efficiency.

\section{Related Work}

Over recent years, there has been a significant upsurge in research efforts dedicated to overcoming the challenges of few-shot learning within specialized areas of medical imaging. This section provides insights into various contributions in this field, highlighting how researchers have employed diverse few-shot learning techniques to address a range of medical diagnostic problems. For instance, Mahajan et al.\cite{skindis} focused on skin disease identification in their study. They explored the effectiveness of two well-known few-shot learning methods, Reptile \cite{Reptile} and Prototypical Networks \cite{Prototypical}, in distinguishing between different skin diseases. Another notable study by Medela et al. \cite{Relsiam} tackled the challenge of transferring knowledge across different tissue types. Their approach involved training a deep Siamese neural network to transfer knowledge from a dataset containing colon tissue to another dataset encompassing colon, lung, and breast tissue. Teng et al. \cite{Rellungcarc} proposed a specialized few-shot learning algorithm based on Prototypical Networks, specifically designed for classifying lymphatic metastasis of lung carcinoma from Whole Slide Images (WSIs). In a different approach, Chen et al. \cite{chestct} adopted a two-stage framework for the critical task of COVID-19 diagnosis from chest CT images. They first captured expressive feature representations using an encoder trained on publicly available lung datasets through contrastive learning. Subsequently, in the second stage, the pre-trained encoder was utilized within a few-shot learning paradigm, leveraging the Prototypical Networks method. Yang et al. \cite{RelCL} introduced an innovative approach by incorporating contrastive learning (CL) with latent augmentation (LA) to build a few-shot classification model. Here, contrastive learning learns significant features without requiring labels, while Latent Augmentation transfers semantic variations from one dataset to another without supervision. The model was trained on publicly available colon cancer datasets and evaluated on PAIP \cite{PAIP} liver cancer Whole Slide Images. Additionally, Shakeri et al. \cite{FHIST} introduced the FHIST dataset along with evaluations of various few-shot classification methodologies, which serve as a catalyst for further exploration in this domain. The scope of few-shot learning extends beyond image classification, as evidenced by \cite{Relseg1}'s introduction of a few-shot semantic segmentation framework to reduce reliance on labeled data during training. Likewise, \cite{Relseg2} explores medical image segmentation using the location-sensitive local prototype network, integrating spatial priors to improve segmentation accuracy.

\section{Methodology}
Few-shot classification tackles the challenge of working with limited data instances, such as in medical datasets where rare classes may only have a handful of examples, typically around five images per class. Training deep learning (DL) models with such a small dataset can be exceedingly difficult and may lead to overfitting. Furthermore, since these classes are rare, it's probable that pre-trained models have not encountered similar images during their training phase. Consequently, traditional transfer learning approaches may not be effective in this scenario, prompting the adoption of few-shot models that follow an episodic-training paradigm.

\begin{figure*}[htbp]
\centering
\includegraphics[height=5cm,width=10cm]{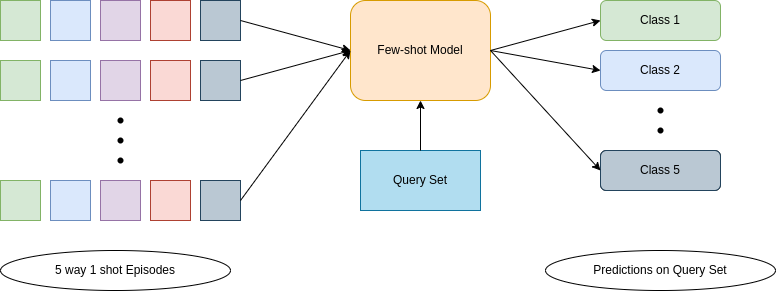}
\caption{The diagram depicts a few-shot learning model, demonstrating its capability to generalize effectively and recognize classes within an unlabeled query set using only a limited number of support examples. In the illustration, five different colors in the support set represent five distinct classes (ways), each having one sample (shot). \cite{fewshotbioimaging}}
\label{fig1}
\end{figure*}

In few-shot training, the training and test sets are disjoint. The training set, denoted as $D_{Train}$, comprises pairs $(X_{Train},Y_{Train})_{i=1}^{M}$, where $X_{Train}$ represents the images and  $Y_{Train}$ denotes the corresponding labels. Here, M denotes the number of classes. In episodic training, the extensive labeled training dataset is divided into multiple episodes. Each episode consists of a support set(S) and a query set(Q). These episodes are characterized by the K-way N-shot Q-query setup. Essentially, for each episode, K classes are randomly chosen from the total M classes. Within each of these K classes, N images are designated for the support set, while Q images are reserved for the query set. Consequently, the support set comprises N$\times$K images, while the query set comprises Q$\times$K images. The few-shot model is then trained to learn from the support set in order to accurately predict the labels of the query set. Following training, the model's performance is evaluated on the test set.

\begin{equation}
D_{Train}=(X_{Train},Y_{Train})
\label{eqn1}
\end{equation}
\begin{equation}
S := {(X_{i},Y_{i})_{i=1}^{K \times N}}
\label{eqn2} 
\end{equation}
\begin{equation}
Q := {(X_{i},Y_{i})_{i=1}^{K \times Q}}
\label{eqn3} 
\end{equation}
\begin{equation}
D_{Test}=(X_{Test},Y_{Test})
\label{eqn4}
\end{equation}
\begin{equation}
S := {(X_{i},Y_{i})_{i=1}^{K \times N}}
\label{eqn} 
\end{equation}
\begin{equation}
Q := {(X_{i})}
\label{eqn} 
\end{equation}
\begin{equation}
Y_{Train} \cap Y_{Test} = \Phi
\label{eqn5}
\end{equation}

\subsection{Prototypical Networks}

The fundamental principle behind Prototypical Networks \cite{Prototypical} lies in the recognition that data points, which exhibit closeness to a singular prototype representation for each class, contribute significantly to a meaningful embedding. To translate this concept into practice, an essential step involves a non-linear mapping that transforms the input data into a specialized embedded space. Within this embedded space, each class's prototype representation is derived by computing the mean of its respective support set. The process commences with the establishment of a non-linear mapping, which is pivotal for capturing intricate relationships and patterns within the data, effectively transforming the original input data into a specialized embedded space. Within this space, a prototype representation is defined for each category, serving as a central point representing the class. This prototype is computed by averaging the feature vectors of all instances in the support set belonging to that class. During classification, the approach relies on identifying the nearest distance between the embedded query and the prototype representation of each class. The class associated with the closest prototype is then assigned to the query, thereby determining its classification. The support set S comprises K labeled examples, represented as S=[($X_1$,$Y_1$),..,($X_K$,$Y_K$)] where  $X_i$ denotes a D-dimensional feature vector for each image and $Y_i$ represents the corresponding label of $X_i$. Within the support set, there are N classes, with the number of examples within each class denoted as $S_N$.

Prototypical Networks employ an embedding function, typically a Convolutional Neural Network (CNN), denoted as 
$f_{\Phi}$, to estimate an M-dimensional representation $C_N$ for each class. Here, $C_N$ represents the mean vector of the support points belonging to each class. The parameters $\Phi$ are learned during the training process, allowing the network to adapt and optimize its embedding function to effectively capture the underlying structure and characteristics of the data.

\begin{equation}
f_\Phi:R^D -> R^M 
\label{eqn6}
\end{equation}
\begin{equation}
C_N \in R^MRellungcarc
\label{eqn7}
\end{equation}
\begin{equation}
C_N=\frac{1}{|S_N|}\sum_{(X_i,Y_i)\in S}f_{\Phi}(X_i)
\label{eqn8}
\end{equation}

Prototypical Networks generate a probability distribution across classes for a given query example X. This distribution is computed by applying a softmax function to the distances between the query example and other prototypes in the embedding space. Specifically, the softmax function normalizes these distances, transforming them into a probability distribution that indicates the likelihood of the query belonging to each class represented by the prototypes.

\begin{equation}
P_{\Phi}(Y=N|X)=\frac{\exp(-d(f_{\Phi}(X),C_N))}{{\sum_{N^{'}}}\exp(-d(f_{\Phi}(X),C_{N^{'}}))}
\label{eqn9}
\end{equation}

The model learns by minimizing the negative log probability $J(\Phi)$ of the true class N using the Adam solver. 

\begin{equation}
J(\Phi)=-\log(P_{\Phi}(Y=N|X))
\label{eqn10}
\end{equation}

\subsection{SimpleShot}

SimpleShot~\cite{Simpleshot} is a straightforward non-episodic few-shot learning approach that leverages transfer learning and the nearest-neighbor rule. Initially, a large-scale deep neural network is trained on a set of training classes, effectively learning rich representations of the data. During testing, the trained deep neural network serves as a feature encoder. The nearest-neighbor rule is then applied to the images of the test episodes using these learned features. This process enables SimpleShot to make predictions for new classes based on their similarity to the learned representations of the training classes, without requiring the episodic training paradigm typically employed in other few-shot learning methods.

The training set, denoted as $D_{Train}$, consists of image-label pairs, where X represents the images and Y denotes the corresponding labels. This training set is used to train a Convolutional Neural Network (CNN) using the cross-entropy loss function. 
In the context where $f_\theta$ represents the Convolutional Neural Network (CNN), W represents the weights of the classification layer, and L represents the cross-entropy loss function, the training process involves optimizing the parameters $\theta$ of the CNN to minimize the loss function L.

\begin{equation}
D_{Train}=[(X_1,Y_1),(X_2,Y_2),...,(X_N,X_N)]
\label{eqn17}
\end{equation}
\begin{equation}
argmin_\theta\sum_{(X,Y)\in D_Train}{l(W^T f_\theta(X),Y)}
\label{eqn18}
\end{equation}

Testing involves utilizing the nearest neighbor rule for classification. Initially, an image's features are extracted by feeding it through the CNN, resulting in feature representation denoted as Z. Subsequently, classification via nearest neighbor rule is executed, employing Euclidean distance calculations between 
Z and the feature representations of test set episodes. 

\begin{equation}
Z=f_\theta(X)
\label{eqn19}
\end{equation}

In a one-shot setting, the support set S of the test set comprises one labeled example from each of the N classes.

\begin{equation}
S=((\hat{X}_1,1),...,(\hat{X}_N,N))
\label{eqn20}
\end{equation}

Utilizing the Euclidean distance measure, the nearest neighbor rule assigns the query image $\hat{X}$ to the class of the most similar support image.

\begin{equation}
Y(\hat{X})=argmin_{N\in(1,2,..,N)}d(\hat{Z},\hat{Z}_N)
\label{eqn21}
\end{equation}

In a multi-shot setting, $\hat{Z}$  and  $\hat{Z}_N$ represent the CNN features of the query and support images, respectively. Specifically,  $\hat{Z}_N$ is computed as the average feature vector of each class in the support set.

\subsection{Laplacianshot}

LaplacianShot\cite{Laplacianshot} presents a novel methodology for addressing few-shot tasks: transductive Laplacian regularized interference. This approach is centered around minimizing a quadratic binary assignment function that consists of two crucial components. The first component, a unary term, is responsible for assigning query samples to their nearest class prototypes. Meanwhile, the second component, the pairwise-Laplacian term, serves to encourage consistent label assignments among neighboring query samples.

Similar to the methodology employed in SimpleShot, the training dataset, denoted as $D_{Train}$, is utilized. In this approach, the training dataset is fed into a convolutional neural network (CNN) and trained using a basic cross-entropy loss function. Notably, this methodology does not incorporate any episodic or meta-learning strategies.

\begin{equation}
argmin_\theta\sum_{(X,Y)\in D_Train}{l(W^T f_\theta(X),Y)}
\label{eqn22}
\end{equation}

Below are the regularization equations utilized during few-shot test inference:

\begin{equation}
E(Y)=N(Y)+\lambda \frac{1}{2} L(Y)
\label{eqn23}
\end{equation}
\begin{equation}
N(Y)=\sum_{q=1}^N \sum_{c=1}^C y_{q, c} d(z_q-m_c)
\label{eqn24}
\end{equation}
\begin{equation}
L(Y)=\frac{1}{2} \sum_{q, p} w(z_q, z_p)||y_q-y_p||_2^2
\label{eqn25}
\end{equation}

In this task, we aim to minimize two components. The first component, denoted as N(Y), is minimized by assigning each query point to the class of its nearest prototype $m_c$ from the support set. This assignment is made using a distance metric like Euclidean distance. The second component, L(Y), serves as the Laplacian regulariser and is given by tr($Y^{T}$LY). Here, L represents the Laplacian matrix, which corresponds to the affinity matrix W =  $w(z_q, z_p)$. . This matrix quantifies the similarity between feature vectors $z_q$ and  $z_p$ using a kernel function. Specifically, $z_q$ and $z_p$ denote the feature vectors of query images  $x_p$ and $x_q$, respectively.

\subsection{DeepEMD}

In DeepEMD\cite{DeepEMD}, the Earth Mover's Distance (EMD) is employed as a metric to compute the structural distance between dense image representations, thereby assessing image similarity. By facilitating optimal matching flows between structural elements, EMD minimizes the matching cost, which in turn serves as an indicator of image distance for classification purposes. To establish crucial weights for elements within the EMD framework, a cross-reference mechanism is introduced. This mechanism helps alleviate the impact of clustered backgrounds and intra-class variations. For K-shot classification tasks, a structured fully connected layer is utilized, enabling direct classification of dense image representations using EMD.

The Earth Mover's Distance functions as a distance metric between two sets of weighted objects or distributions, drawing upon the core concept of distance between individual objects. It mirrors the structure of the Transportation Problem within Linear Programming. In this problem, a set of suppliers S = ({$S_i$$|$i=1,2....m}) is tasked with transporting goods to a set of demanders ({$d_j$$|$j=1,2....k}). Here,  $S_{i}$ denotes the supply unit of supplier i, while $d_{j}$ indicates the demand of demander j. The cost of transporting one unit from supplier i to demander j is denoted by $c_{ij}$, and the quantity of units transported is represented by $x_{ij}$.  The primary objective of the transportation problem is to determine the least-expensive flow of goods $\tilde{x}$ = ($\tilde{x}_{ij}$$|$i=-1,2,...,m,j=1,2....,k) from each supplier to each demander.

\begin{equation}
\underset{x_{i j}}{\operatorname{minimize}} \quad \sum_{i=1}^m \sum_{j=1}^k c_{i j} x_{i j}
\label{eqn26}
\end{equation}

In few-shot classification, determining the similarity between support and query images involves passing the images through a fully convolutional network (FCN) to generate image features for both the support set (denoted as S) and the query set (denoted as Q).

\begin{equation}
\mathbf{S} \in \mathbb{R}^{H \times W \times C}
\label{eqn27}
\end{equation}
\begin{equation}
\mathbf{Q} \in \mathbb{R}^{H \times W \times C}
\label{eqn28}
\end{equation}

The matching cost between the two sets of vectors is indicative of the similarity between the images. The cost between the two embeddings $s_{i}$ and $q_{j}$ is determined by:

\begin{equation}
c_{i j}=1-\frac{\mathbf{s}_i^T \mathbf{q}_j}{\left\|\mathbf{s}_i\right\|\left\|\mathbf{q}_j\right\|},
\label{eqn29}
\end{equation}

In a support set, when the shot is greater than 1, the learnable embedding consists of a group of image features for each class rather than a single vector. Subsequently, the mean of these features is computed to obtain a single image feature. This process is akin to calculating the prototype of a class within the support set. The fully connected network serves as a feature extractor, and the Stochastic Gradient Descent (SGD) optimizer is employed to update the weights by sampling few-shot episodes from the dataset.

\subsection{DeepBDC}

In Standard few-shot learning, the process is conducted episodically across various tasks. Each task is typically structured as an N-way K-shot classification challenge, where N represents the number of classes, K denotes the number of support images per class, and Q represents the number of query images per class. These tasks are performed on a support set and a query set. During training, a learner is trained on the support set , learning to classify images into their respective classes. Following training, the learner is evaluated by making predictions on the query set, determining its ability to generalize and classify unseen images accurately.

In DeepBDC\cite{DeepBDC}, the model learns a metric space where classification is achieved by computing distances to the prototype of each class. For a given task (with support set $D_{\text {sup }}$ and query set $D_{\text {que }}$), we feed each image $z_j$ to the network to generate the Bilateral Distance Covariance (BDC) matrix $A_\theta\left(z_j\right)$.
\begin{equation}
\mathbf{P}_k=\frac{1}{K} \sum_{\left(\mathbf{z}_j, y_j\right) \in \mathcal{S}_k} \mathbf{A}_{\boldsymbol{\theta}}\left(\mathbf{z}_j\right)
\end{equation}
$S_k$ represent the set of examples in the support set $D_{\text {sup }}$ labeled with class k. To generate a distribution over classes, a softmax function is computed over the distances from the query examples to the prototypes of the support classes. This softmax operation assigns probabilities to each class based on their similarity to the query examples. The loss function is formulated as the following:
\begin{equation}
\arg \min _{\boldsymbol{\theta}}-\sum_{\left(\mathbf{z}_j, y_j\right) \in \mathcal{D} \text { que }} \log \frac{\exp \left(\tau \operatorname{tr}\left(\mathbf{A}_{\ominus}\left(\mathbf{z}_j\right)^T \mathbf{P}_{y_j}\right)\right)}{\sum_k \exp \left(\tau \operatorname{tr}\left(\mathbf{A}_{\boldsymbol{\theta}}\left(\mathbf{z}_j\right)^T \mathbf{P}_k\right)\right)}
\end{equation}

\section{Experiments and Results}

\begin{figure*}[htbp]
\centering
\includegraphics[height=8cm,width=12cm]{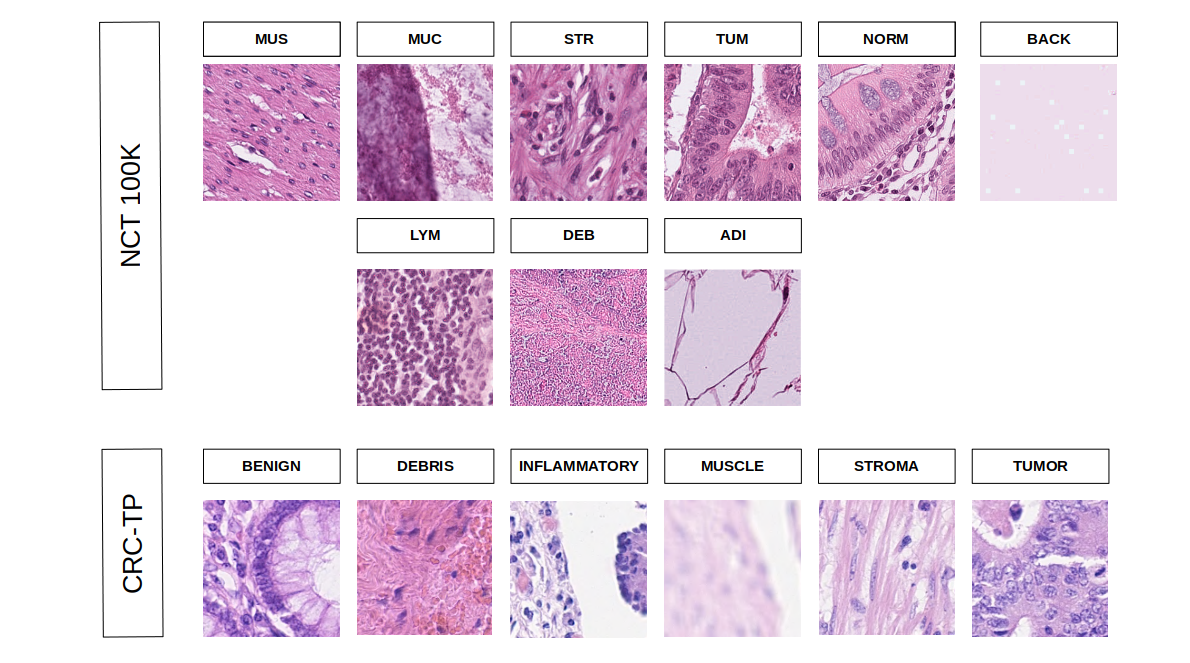}
\caption{The NCT dataset is illustrated by the sample images in the top two rows, whereas the CRC-TP dataset is represented by the sample images in the bottom row. \cite{fewshotbioimaging}}
\label{fig2}
\end{figure*}

\begin{figure*}[htbp]
\centering
\includegraphics[height=8cm,width=12cm]{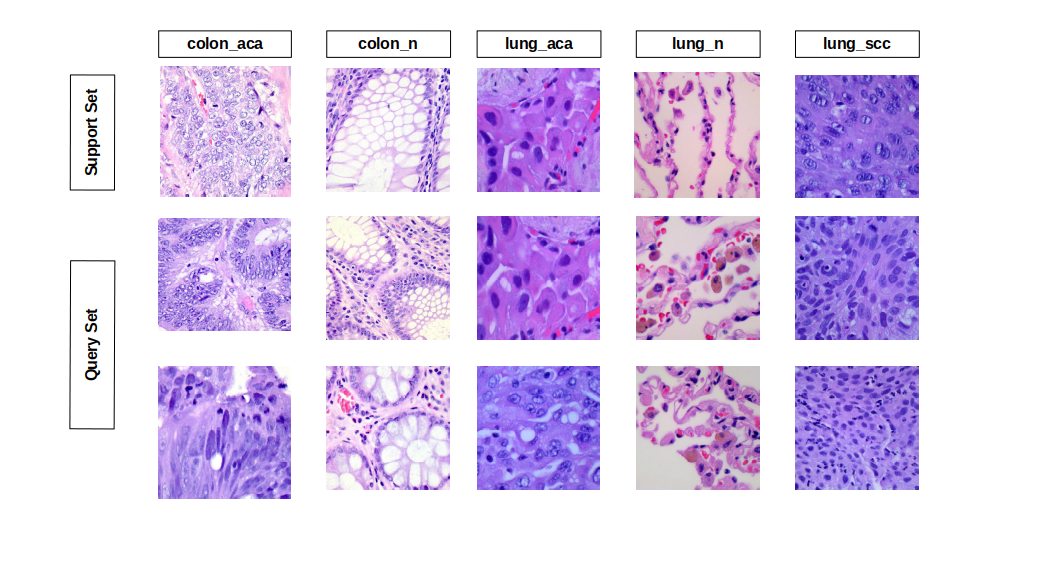}
\caption{5-way 1-shot 2-query episode with support set images from five LC25000 dataset classes in the first row and query set images in the last two rows. \cite{fewshotbioimaging} }
\label{fig3}
\end{figure*}
\subsection{Dataset}

The dataset created by Komura et al. \cite{1.6M} comprises histology images extracted from uniform tumor regions in The Cancer Genome Atlas Program whole slide images \cite{tcga}. The TCGA dataset encompasses tissue slides from 32 cancer types across various human body sites. From this extensive dataset, Komura et al. generated a collection of 1,608,060 image patches, encompassing six different magnification levels ranging from 0.5$\mu$/pixel to 1.0$\mu$/pixel. Given that these images originate from 32 distinct cancer types, they are categorized into corresponding 32 classes. The Mini-Imagenet dataset, proposed by Vinyals et al.\cite{miniimagenet}, comprises of 64,000 images, evenly distributed among 100 classes. Each image in mini-Imagenet has a resolution of 84$\times$84. The NCT dataset \cite{NCT} consists of 100,000 image patches of human colorectal cancer, extracted from Hematoxylin and Eosin stained histological images, alongside normal tissue samples. These images, with a resolution of 224$\times$224, are classified into seven classes: Adipose (ADI), background (BACK), debris (DEB), lymphocytes (LYM), mucus (MUC), smooth muscle (MUS), normal colon mucosa (NORM), cancer-associated stroma (STR), and colorectal adenocarcinoma epithelium (TUM). LC25000 \cite{lc25000}, or the Lung and Colon Histopathological dataset, consists of 25,000 image patches. As the name implies, it includes images from both lung and colon cancer. With a resolution of 768$\times$768, these images are categorized into five classes, three of which belong to lung cancer and the remaining two to colon cancer. The classes include benign colon tissues, colon adenocarcinoma, lung squamous cell carcinoma, and benign lung tissues. CRC-TP \cite{crctp} is yet another colon cancer dataset, comprising 280,000 image patches categorized into six classes: tumor, stroma, complex stroma muscle, debris, inflammatory, and benign. These images have dimensions of 150$\times$150. Mini-Imagenet dataset and the dataset prepared by Komura et al. serves as the training dataset due to their substantial size, suggesting the potential for robust few-shot backbone network training. Conversely, the remaining datasets are utilized as the testing set, adhering to the standard practice in few-shot learning where the test and training sets are disjoint.

\subsection{Results and Discussion}

The experiments were conducted on an NVIDIA A100 using PyTorch, employing two distinct training approaches: episodic training and standard training. Standard training involves training across the entire dataset with ample examples per class, while episodic training, tailored for few-shot learning, enables the model to quickly generalize from small episodes containing minimal examples of new classes. Methods like Protonet, DeepEMD, and DeepBDC adopt episodic training, whereas SimpleShot and LaplacianShot adhere to standard training procedures.

During episodic training, the methods undergo training for 120 epochs. Each epoch involves selecting 600 random episodes from the training set. The models are trained separately for 5-way 1-shot, 5-way 5-shot, and 5-way 10-shot scenarios, with the number of query images set to 15 in all cases. The initial learning rate is set to 1e-3, with a $\gamma$ value of 0.1. The batch size is configured to accommodate one episode, with the number of images in a batch calculated as 80, 100, and 125 for 5-way 1-shot, 5-way 5-shot, and 5-way 10-shot training, respectively. Conversely, few-shot methods employing standard CNN training are trained for 150 epochs, with a batch size of 512. The initial learning rate and weight decay are initialized to 0.05 and 5e-4, respectively. ResNet18 serves as the backbone network in both episodic and standard training procedures.
All trained models undergo meta-testing, wherein 5000 episodes are randomly sampled from the CRC-TP, NCT, and LC25000 datasets. Regardless of the K-shot testing scenario, the number of query images remains fixed at 15 in each episode. The results on 1-shot, 5-shot, 10-shot models on different datasets trained on dataset proposed by Komura et al.\cite{1.6M} and Mini-Imagenet dataset\cite{miniimagenet} are reported in the Table 1 and Table 2 respectively.

\begin{table*}
\caption{Accuracy(\%) on three different datasets; CRC-TP \cite{crctp}, NCT \cite{NCT} and LC25000 \cite{lc25000}. Komura et al. \cite{1.6M} proposed the dataset used to train the few-shot models. \cite{fewshotbioimaging}}
\begin{center}
\begin{tabular}{lcccc}
\hline  & \multicolumn{3}{c}{ CRC-TP } \\
Method &  Training Method &  5-way 1-shot & 5-way 5-shot & 5-way 10-shot \\
\hline ProtoNet \cite{Prototypical} & Episodic & $43.8$ & $63.6$ & $68.3$ \\
DeepEMD \cite{DeepEMD} & Episodic & $47.3$ & $64.6$ & $68.6$ \\
DeepBDC \cite{DeepBDC} & Episodic & $47.7$ & $65.3$ & $70.2$ \\
SimpleShot \cite{Simpleshot} & Standard & $47.9$ & $66.9$ & $71.4$ \\
\textbf{LaplacianShot} \cite{Laplacianshot} & \textbf{Standard} & $\mathbf{48.5}$ & $\mathbf{68.0}$& $\mathbf{72.8}$ \\
\hline & &  NCT \\
Method & Training Method & 5-way 1-shot & 5-way 5-shot & 5-way 10-shot \\
\hline ProtoNet \cite{Prototypical} & Episodic & $62.6$ & $80.9$ & $84.9$ \\
DeepEMD \cite{DeepEMD} & Episodic & $68.5$ & $84.0$ & $86.0$ \\
DeepBDC \cite{DeepBDC}  & Episodic & $69.3$ & $84.7$ & $87.5$ \\
SimpleShot \cite{Simpleshot} & Standard & $71.2$ & $85.5$ & $88.2$ \\
\textbf{LaplacianShot} \cite{Laplacianshot} & \textbf{Standard} & $\mathbf{71.8}$ & $\mathbf{86.9}$ & $\mathbf{89.5}$ \\
\hline &  \multicolumn{3}{c}{ LC25000 } \\
Method & Training Method & 5-way 1-shot & 5way 5-shot & 5-way 10-shot \\
\hline ProtoNet \cite{Prototypical}  & Episodic & $67.2$ & $84.8$ & $86.2$ \\
DeepEMD \cite{DeepEMD}  & Episodic & $73.8$ & $85.3$ & $86.4$ \\
DeepBDC \cite{DeepBDC}  &Episodic& $74.7$ & $85.8$ & $86.9$ \\
SimpleShot \cite{Simpleshot}  & Standard & $66.4$ & $83.6$ & $87.2$ \\
\textbf{LaplacianShot} \cite{Laplacianshot}  & \textbf{Standard} & $\mathbf{67.5}$ & $\mathbf{84.2}$ & $\mathbf{87.9}$ \\
\hline\
\end{tabular}
\end{center}
\end{table*}

\begin{figure}[htbp]
\includegraphics[height=8cm,width=12cm]{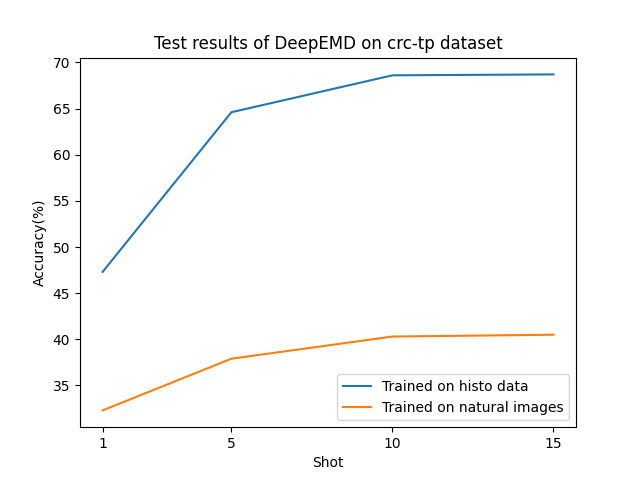}
\caption{Performance of DeepEMD on crc-tp dataset with different training conditions}
\label{fig4}
\end{figure}
\begin{figure}[htbp]
\includegraphics[height=8cm,width=12cm]{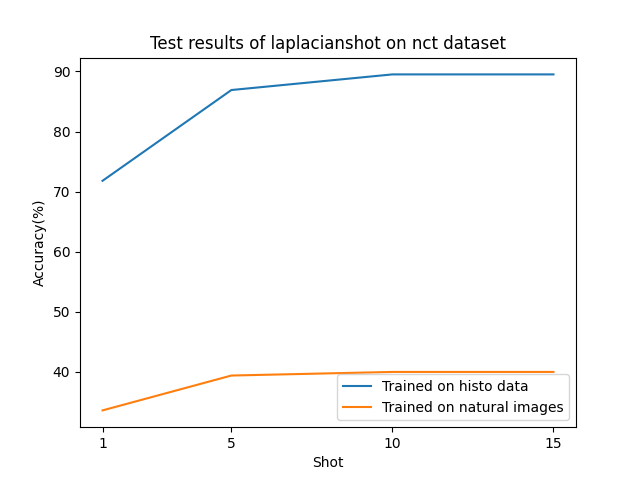}
\caption{Performance of laplacianshot on nct dataset with different training conditions}
\label{fig5}
\end{figure}

When trained on Mini-Imagenet i.e. natural images' dataset, the few-shot models did not exhibit promising performance. On testing these models on histopathology data, their accuracies and general capabilities are significantly low. This suggests that the feature encoders learned from natural images dataset might not adequately capture the unique characteristics and complexities present in histopathology images. Conversely, when trained directly on histopathology data, the same few-shot classification models demonstrated significantly improved performance across all tested scenarios (5-way 1-shot, 5-way 5-shot, and 5-way 10-shot). The enhanced results indicate that training the models on domain-specific data enables them to learn feature representations that are better suited for histopathology image classification tasks. By directly optimizing the feature encoder on histopathology data, the models can effectively capture the intricate patterns and subtle nuances characteristic of histopathological images, leading to improved classification accuracy and robustness. As expected, few-shot methodologies employing standard fine-tuning methods like SimpleShot and LaplacianShot demonstrated superior performance when compared to episodic training approaches such as ProtoNet, DeepEMD, and DeepBDC. This could be attributed to the fact that the standard training techniques allowed these models to leverage the extensive dataset provided by Komura et al., encompassing cancer image patches from 32 distinct organ classes. In 10-shot test scenarios, the accuracies of all methods across all datasets, except for CRC-TP, ranged from 85\% to 90\%.

\begin{table*}
\caption{Table 2: Accuracy(\%) on three different datasets; CRC-TP \cite{crctp}, NCT \cite{NCT} and LC25000 \cite{lc25000}. The few-shot models are trained on Mini-Imagenet \cite{miniimagenet}}
\begin{center}
\begin{tabular}{lcccc}
\hline  & \multicolumn{3}{c}{ CRC-TP } \\
Method &  Training Method &  5-way 1-shot & 5-way 5-shot & 5-way 10-shot \\
\hline 
ProtoNet \cite{Prototypical} & Episodic & $28.9$ & $36.3$ & $39.3$ \\
DeepEMD \cite{DeepEMD} & Episodic & $32.3$ & $37.9$ & $40.3$ \\
DeepBDC \cite{DeepBDC} & Episodic & $31.9$ & $36.5$ & $41.2$ \\
SimpleShot \cite{Simpleshot} & Standard & $32.4$ & $\mathbf{44.7}$ & $\mathbf{49.3}$ \\
{LaplacianShot} \cite{Laplacianshot} & {Standard} & $\mathbf{34.0}$ & $39.3$& $40.4$ \\
\hline & &  NCT \\
Method & Training Method & 5-way 1-shot & 5-way 5-shot & 5-way 10-shot \\
\hline ProtoNet \cite{Prototypical} & Episodic & $26.3$ & $33.9$ & $37.4$ \\
DeepEMD \cite{DeepEMD} & Episodic & $28.9$ & $34.6$ & $38.5$ \\
DeepBDC \cite{DeepBDC}  & Episodic & $30.1$ & $34.5$ & $39.8$ \\
SimpleShot \cite{Simpleshot} & Standard & $32.3$ & $\mathbf{44.9}$ & $\mathbf{49.1}$ \\
\textbf{LaplacianShot} \cite{Laplacianshot} & Standard & $\mathbf{33.6}$ & $39.4$ & $40.0$ \\
\hline &  \multicolumn{3}{c}{ LC25000 } \\
Method & Training Method & 5-way 1-shot & 5way 5-shot & 5-way 10-shot \\
\hline ProtoNet \cite{Prototypical} & Episodic & $30.2$ & $35.5$ & $41.1$ \\
DeepEMD \cite{DeepEMD} & Episodic & $29.5$ & $35.9$ & $39.7$ \\
DeepBDC \cite{DeepBDC} & Episodic & $31.3$ & $35.7$ & $40.7$ \\
SimpleShot \cite{Simpleshot}  & Standard & $32.5$ & $\mathbf{44.8}$ & $\mathbf{49.2}$ \\
LaplacianShot \cite{Laplacianshot}  & Standard & $\mathbf{33.9}$ & $39.2$ & $40.1$ \\
\hline\
\end{tabular}
\end{center}
\end{table*}

\section{Conclusion}

 Upon conducting rigorous experiments with a range of few-shot classification techniques, including SimpleShot, LaplacianShot, Prototypical Networks, DeepEMD, and DeepBDC, across different training and testing datasets, it became apparent that training these models on histopathology data resulted in significantly improved performance compared to training on natural images. This notable improvement underscores the critical role of obtaining a specialized feature encoder tailored specifically for histopathology data. Histopathology images possess unique characteristics and complexities that differ from natural images, such as varied tissue structures, cellular compositions, and staining patterns. By training the models on histopathology data, the feature encoder becomes adept at capturing and extracting relevant discriminative features inherent to histopathological images, leading to enhanced classification accuracy and robustness.  Overall, these findings underscore the importance of training few-shot classification models on domain-specific data to obtain a good feature encoder tailored to the target application domain.  Therefore, investing efforts in developing and optimizing feature encoders specifically designed for histopathology data is essential for achieving superior performance in few-shot classification tasks within the domain of medical imaging.

%
%
%
%

\end{document}